\documentclass[sigconf]{acmart}

\usepackage[utf8]{inputenc}
\usepackage{enumitem}
\usepackage{amsmath}
\usepackage{multirow}

\title{Local Connectivity in Centroid Clustering}
\author{Deepak P}
\affiliation{%
  \institution{Queen's University Belfast, UK}
}
\email{deepaksp@acm.org}

\copyrightyear{2020}
\acmYear{2020}
\setcopyright{acmlicensed}\acmConference[IDEAS 2020]{24th International Database Engineering & Applications Symposium}{August 12--14, 2020}{Seoul, Republic of Korea}
\acmBooktitle{24th International Database Engineering & Applications Symposium (IDEAS 2020), August 12--14, 2020, Seoul, Republic of Korea}
\acmPrice{15.00}
\acmDOI{10.1145/3410566.3410601}
\acmISBN{978-1-4503-7503-0/20/06}

\begin{document}

\begin{abstract}
Clustering is a fundamental task in unsupervised learning, one that targets to group a dataset into clusters of similar objects. There has been recent interest in embedding normative considerations around fairness within clustering formulations. In this paper, we propose 'local connectivity' as a crucial factor in assessing membership desert in centroid clustering. We use local connectivity to refer to the support offered by the local neighborhood of an object towards supporting its membership to the cluster in question. We motivate the need to consider local connectivity of objects in cluster assignment, and provide ways to quantify local connectivity in a given clustering. We then exploit concepts from density-based clustering and devise LOFKM, a clustering method that seeks to deepen local connectivity in clustering outputs, while staying within the framework of centroid clustering. Through an empirical evaluation over real-world datasets, we illustrate that LOFKM achieves notable improvements in local connectivity at reasonable costs to clustering quality, illustrating the effectiveness of the method. 
\end{abstract}



\maketitle

\section{Introduction}

Clustering~\cite{jain1999data} has been a popular task in unsupervised learning. Clustering involves grouping a dataset of objects into a number of groups such that objects that are highly similar to one another are more likely to find themselves assigned to the same group, and vice versa. Clustering algorithms fall into one of many families, of which partitional and hierarchical algorithms are two main streams. Partitional clustering, arguably the more popular stream, considers grouping the dataset into a number of disjoint sets. The pioneering work in this family, $K$-Means clustering, dates back to the 1960s~\cite{macqueen1967some}. $K$-Means clustering is a partitional clustering algorithm that additionally outputs a prototypical object to {\it 'represent'} each cluster, which happens to simply be the cluster {\it centroid} within the basic $K$-Means formulation. The centroid output is often seen as very useful for scenarios such as for manual perusal to ascertain cluster characteristics, resulting in this paradigm of {\it 'centroid clustering'}~\cite{taillard2003heuristic} attracting much research interest. In alternative formulations within the centroid clustering paradigm, the prototypical object is set to be the medoid, which is a dataset object that is most centrally positioned; this is referred to as $K$-medoids~\cite{rdusseeun1987clustering} clustering or PAM\footnote{https://en.wikipedia.org/wiki/K-medoids}. 50+ years since $K$-Means, the basic $K$-Means formulation is still used widely and continues to inspire much clustering research~\cite{jain2010data}. The second popular family of clustering algorithms, that of hierarchical clustering, focuses on generating a hierarchy of clusters from which clusterings of differing granularities can be extracted. An early survey of hierarchical clustering methods appears at~\cite{murtagh1983survey}. Our focus in this paper is within the task of centroid clustering. 

\subsection{Membership Desert in Centroid Clustering}\label{sec:introdesert}

In this paper, we problematize the notion of {\it cluster membership} in centroid clustering from a conceptual and normative perspective. Our work is situated within the context of recent interest in fairness and ethics in machine learning (e.g.,~\cite{loi2019include}), which focuses on embedding normative principles within data science algorithms in order to align them better with values in the modern society. In particular, we consider the question of {\it membership desert}, or what it means for an object to be deserving of being a member of a cluster, or a cluster to be deserving of containing a data object. Desert in philosophical literature\footnote{https://en.wikipedia.org/wiki/Desert\_(philosophy)} refers to the condition of being deserving of something; a detailed exposition of philosophical debate on the topic can be found within a topical encyclopaedia from Stanford\footnote{https://plato.stanford.edu/entries/desert/}. $K$-Means and most other formulations that build upon it have used a fairly simple notion of membership desert; that an object be assigned to the cluster to whose prototype it is most proximal, according to a task-relevant notion of similarity. While this simple notion makes intuitive sense as well as enables convenient optimization, it admits unintuitive outcomes as we will see later. 

There have been two recent works in re-considering membership desert in centroid clustering, both within the umbrella of research in fair machine learning. The first work~\cite{chen2019proportionally} considers a notion of {\it collective desert} to blend in with the $K$-Means framework, whereby a reasonably large set of objects is considered to be deserving of their own cluster as long as they are collectively proximal to one another. The second work~\cite{repfairness} considers the distance-to-centroid as a cost of abstraction incurred by objects in the dataset, and strives to achieve a fair distribution of the cost of abstraction across objects. We will discuss these in detail in a later section. In this work, we consider advancing a third distinct normative consideration in membership desert, that of {\it local connectivity}. At the high level, we consider the membership desert associated with an object-cluster pair as being intimately related to the extent of the object's neighbors' affinity towards the cluster in question. 

\subsection{Our Contributions}

In what may be seen as a contrast to conventional research narratives within data analytics, our work is centered on advancing a particular normative consideration as opposed to a technological challenge. This is in line with recent work on fairness and ethics in AI, which have mostly appeared within data analytics avenues as well (e.g.,~\cite{abraham2020fairness,chen2019proportionally,bera2019fair}). Our contribution by way of this work is three-fold:

\begin{itemize}[leftmargin=*]
    \item {\bf Local Connectivity as Membership Desert:} We develop an argument for considering {\it local connectivity} as a notion of membership desert in centroid clustering. Building upon this argument, we develop quantitative metrics to evaluate the extent to which local connectivity is being adhered to, within a clustering. 
    \item {\bf LOFKM:} We develop a simple centroid clustering formulation, {\it LOFKM}, drawing inspiration from both centroid clustering and density-based clustering, that deepens local connectivity in clustering outputs. 
    \item {\bf Evaluation:} Through an empirical evaluation over multiple real-world datasets, we illustrate that LOFKM is able to significantly improve alignment with local connectivity considerations at reasonable costs to clustering quality. 
\end{itemize}


\noindent{\bf Roadmap:} We start by considering related work in Section~\ref{sec:relwork}, followed by an overview of membership desert in Section~\ref{sec:background}. This is followed by Section~\ref{sec:locconn} where we describe local connectivity as a distinct notion of membership desert and ways of quantifying it for a given clustering. Section~\ref{sec:lofkm} outlines a simple method for enhancing local connectivity in centroid clustering, codenamed $LOFKM$. This is followed by our experimental evaluation in Section~\ref{sec:expts}, a brief discussion in Section~\ref{sec:discussion} and conclusions in Section~\ref{sec:conclusions}. 

\section{Related Work}\label{sec:relwork}

Given that our work advances a local neighborhood based normative consideration in clustering, we briefly summarize related work from (i) fair clustering, and (ii) local neighborhood estimations from the density-based clustering family. 

\begin{figure*}[t]
  \includegraphics[width=0.6\textwidth]{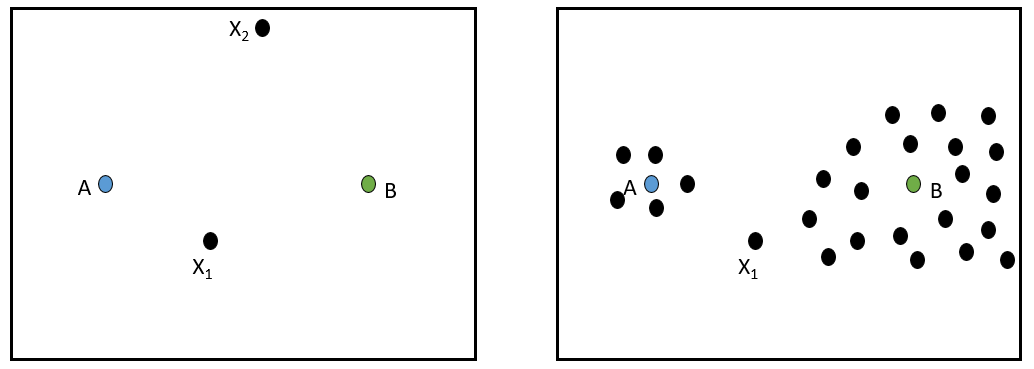}
  \caption{Two Cases for Section~\ref{sec:critiquekmeans}: Rough Illustration}
  \label{fig:kmcritique}
\end{figure*}

\subsection{Fair Clustering}

There has been an emerging interest in fair clustering. Among the two notions of fairness, {\it individual} and {\it group} fairness~\cite{binns2020apparent}, fair clustering has largely seen explorations on the latter. Group fairness involves ensuring cluster-level representational parity of sensitive groups defined on attributes such as gender, race, ethnicity and marital status. This literature, initiated by a work on ingenious dataset pre-processing~\cite{chierichetti2017fair}, has seen work on embedding fairness within the optimization~\cite{abraham2020fairness} as well as in post-processing~\cite{bera2019fair}. These also differ in the number of types of sensitive attributes that they admit. An overview of recent work on group-fair clustering appears in~\cite{abraham2020fairness} (Ref. Table 1 therein). Research into individual fairness in clustering has a flavour of considering membership desert as the focus question; being pertinent to our work, we discuss this in detail in Section~\ref{sec:background}. 

\subsection{Local Neighborhood and Clustering}\label{sec:localneighborhood}

Local neighborhood of objects has been the core consideration in work on density-based clustering, a field pioneered by the DBSCAN clustering algorithm~\cite{ester1996density}, followed by OPTICS~\cite{ankerst1999optics}. In our work, we will make use of a work that extends concepts from density-based clustering in order to identify the outlierness of dataset objects, called {\it Local Outlier Factor} (LOF)~\cite{breunig2000lof}. The structure of LOF relies on quantifying the {\it local density} around an object. The local density around an object is inversely related to the average {\it reachability} of the object to its $k$ nearest neighbors; with {\it reachability} being a lower-bounded version of distance between the objects. The local density around an object's neighbors is then contrasted with the object's own local density to arrive at the LOF, which is a non-negative real number. $LOF>1$ ($LOF<1$) is achieved by objects whose neighbors are in neighborhoods that are denser (sparser) than it's own, with $LOF=1$ indicating a good match between respective densities. Objects with high values of $LOF$, especially $LOF >> 1$, are considered density-based outliers, due to their (relative) lack of closeby neighbors. Over the past two decades, LOF has evolved to being a very popular outlier detection method, continuously inspiring systems work on improving efficiency (e.g.,a recent {\it fast LOF} work appears in~\cite{babaei2019detecting}), arguably adorning a place in the outlier detection literature only next to the analogous status of $K$-Means within clustering literature. 

\section{Background: Membership Desert in Centroid Clustering}\label{sec:background}

Following up from Section~\ref{sec:introdesert}, we now cover more background on the notion of membership desert in $K$-Means, and recent fairness oriented re-considerations of the notion. 

\subsection{Critiquing K-Means' Membership Desert}\label{sec:critiquekmeans}

Let us start with looking at the simple notion of membership desert used in $K$-Means, that an object deserves to be assigned to the cluster whose prototype\footnote{we use prototype and centroid interchangeably} it is most proximal to, proximity measured under a domain-specific notion of (dis)similarity that is deemed relevant to the clustering task. {\it First}, consider the case of two clusters, $A$ and $B$. Now, let an object $X_1$ be at a distance of $3$ and $5$ units from the prototypes of $A$ and $B$ respectively, as shown roughly in the first illustration in Fig~\ref{fig:kmcritique}. For another object $X_2$, also shown in the illustration, let the distances be $8$ and $6$ respectively. The simple $K$-Means ({\it argmin}) heuristic does the following assignment: $X_1 \in A$ and $X_2 \in B$. It may be noted that while considering proximity as membership desert as in $K$-Means, $X_1$ may be considered more deserving of being assigned to $B$ than $X_2$ is to $B$; this is so since $dist(X_1,B)<dist(X_2,B)$. However, the $K$-Means assignment is in conflict with this observation, due to the higher degree of proximity of $X_1$ to $A$. {\it Second}, consider a scenario with respect to the trio, $X_1$ in relation to $A$ and $B$, as shown in the right-side in Figure~\ref{fig:kmcritique}. Let $B$ be a naturally bigger and denser cluster with significant number of data objects within $6$ units of distance of it. On the other hand, let $A$ be a small cluster with most of its members being within $2$ units of distance around its prototype. In this setting, despite $dist(X_1,A) < dist(X_1,B)$, $X_1$ may be thought of as deserving of being located within $B$ since it is in the company of the large mass of points stretching to the proximity of $B$. This intuitive notion of membership desert also conflicts with the cluster assignment that $K$-Means does. In fact, this is also an fallout of a fundamental design assumption in $K$-Means, that clusters be modelled as being modeled as Voronoi cells. {\it While we do *not* argue that the $K$-Means choice is inferior to an alternative available choice, it may be seen that there are intuitive opportunities to critique the simple membership desert mechanism in $K$-Means, and that the choice of most proximal centroid is not the only natural choice.} It is also noteworthy that membership assignment is not a final $K$-Means step, making it not entirely appropriate to consider it in isolation as we have done so far. The cluster assignment step is interleaved with the centroid learning step, leading to an interplay of effects of each other.

\subsection{Fairness-orientated Notions of Membership}

As outlined earlier, there are two recent papers, that motivate different considerations in cluster membership assignment. 

\subsubsection{Proportionality~\cite{chen2019proportionally} or Collective Desert in Cluster Membership} 

$K$-Means uses a parameter, the number of expected clusters in the output, commonly denoted as $K$. Thus, on an average, there are $(n/K)$ objects in a $K$-Means cluster. Proportionality, a concept the authors propose, is the notion that if one can find a set of $\lceil n/K \rceil$ data objects that collectively prefer the same candidate centroid in lieu of their current assignments (which involve different centroids/clusters), they deserve a cluster of their own centered at the candidate centroid that they collectively prefer. A clustering would be regarded as violating proportionality if it involves denying this set of $\lceil n/K \rceil$ objects their own cluster that they deserve. They develop algorithms that generate {\it proportionally fair} clusterings, those that do not violate proportionality. 

\subsubsection{Representativity Fairness~\cite{repfairness}} 

A recent work considers human-in-the-loop analytics pipelines where each cluster centroid is perused in order to arrive at a single decision for all objects in the cluster. Within such pipelines and even more generally, objects that are far away from their assigned cluster centroids suffer a higher {\it 'representativity cost'} from the cluster-level abstraction of the dataset provided by the clustering. RFKM, the proposed method, seeks to level off this object-level cost across the objects in the dataset, and move towards what is called {\it representativity fairness}. Operationally, it considers re-engineering the $K$-Means steps in a way that chances of proximity violations such as those in the first example in Section~\ref{sec:critiquekmeans} are reduced. 

\begin{figure*}
  \includegraphics[width=0.6\textwidth]{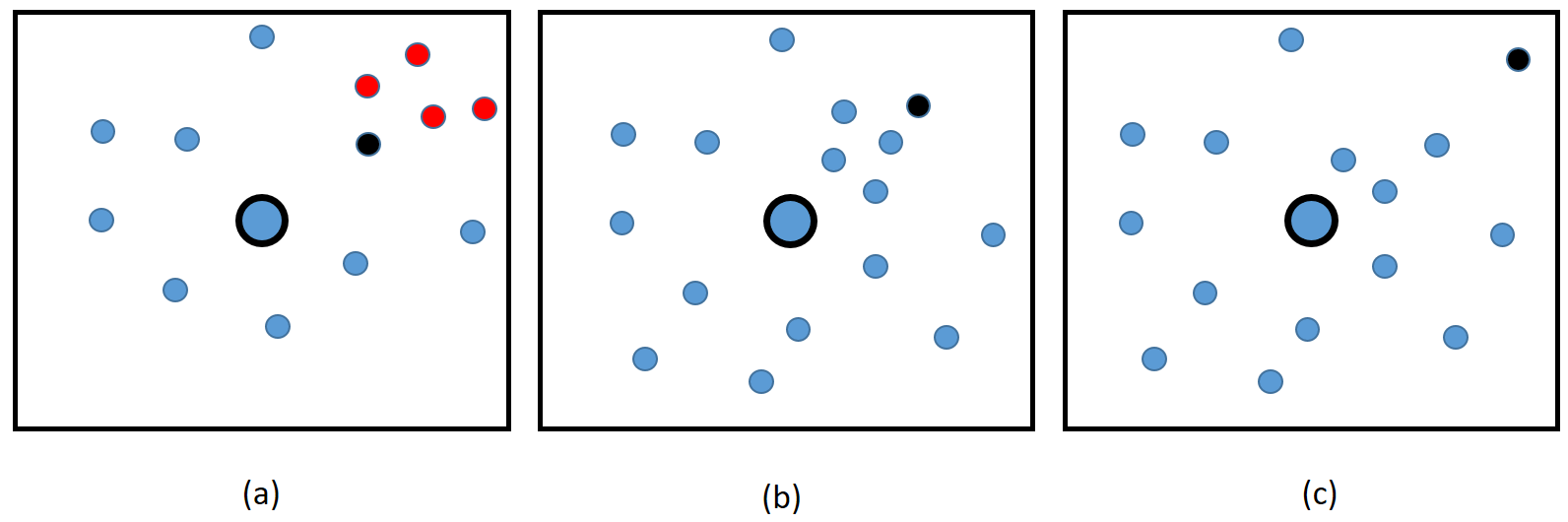}
  \caption{Local Connectivity: Motivating Scenarios (best viewed in color)}
  \label{fig:motivation}
\end{figure*}

\section{Local Connectivity and Membership Desert}\label{sec:locconn}

\subsection{Motivation}

We first consider {\it local connectivity} as a concept and its relevance to membership desert in centroid clustering. Consider three motivating scenarios in Fig.~\ref{fig:motivation}. In each of these figures, the middle point is the designated cluster prototype for the blue cluster; in other words, we have zoomed in on the blue cluster prototype and excluded other points in the dataset (including those from blue or other clusters) from view. The other blue colored points are assigned to be part of the blue cluster, and the red colored points in Fig.~\ref{fig:motivation}(a) are part of a different (red) cluster. In each of these figures, we would like to consider the status of the black colored object, and how well it deserves to be part of the blue cluster, and thus to being {\it 'represented'} by the blue cluster's prototype in the cluster-level abstraction.

Fig~\ref{fig:motivation}(a) has the corresponding black object being closest to the blue cluster prototype among all three scenarios. However, its local neighborhood (think of it as perhaps the closest few data objects to itself) is largely from the red cluster. Intuitively, this makes it reasonable to argue that despite the proximity, the black object in Fig~\ref{fig:motivation}(a) is limited in how well it deserves to be part of the blue cluster; in other words, its membership desert to the blue cluster comes under question. Now, consider the scenario in Fig~\ref{fig:motivation}(b). The black object, while not as proximal as in the case of Fig.~\ref{fig:motivation}(a), is quite well connected to the blue cluster given that it has an {\it 'pull'} from its local neighborhood towards the blue cluster. This makes it more deserving of membership to the blue cluster. Lastly, consider Fig~\ref{fig:motivation}(c) where the black object is tucked into a corner within a sparse region of the space. It has a reasonable claim to membership in the blue cluster, due to its nearest neighbors being blue (despite them being quite far from itself); however, the strength of the claim is dented by its distance to the blue cluster prototype. In summary, we observe the following:

\begin{itemize}[leftmargin=*]
    \item Fig~\ref{fig:motivation}(a): Despite proximity, the membership desert of the black object to the blue cluster is limited due to the local neighborhood being red. 
    \item Fig~\ref{fig:motivation}(b): The black object is most deserving to be part of the blue cluster due to high local connectivity within the blue cluster and reasonable proximity to the blue cluster prototype. 
    \item Fig~\ref{fig:motivation}(c): The black object may be considered as reasonably deserving of blue cluster membership, even though its distance from the blue cluster prototype reduces the strength of the claim. 
\end{itemize}

In other words, these illustrative scenarios offer different trade-offs between the pull towards the blue cluster prototype offered by {\it local connectivity} and {\it proximity}. These, we hope, illustrates that local neighborhood connectivity to the cluster in question is a fairly crucial factor in assessing membership desert. Though we have used abstract examples to motivate local connectivity, this has real-world implications wherever clustering is used for consequential tasks; for a simple example, consider centroid clustering being used for {\it facility location} to determine locations of service facilities (e.g., post offices or hospitals) with people represented using their geographic co-ordinates. In facility location, assigning a person to a facility (located at a centroid) towards which she has few local neighbors may be seen as unjust as well as a decision that undermines social solidarity. 

While $K$-Means is evidently not directly accommodative of local connectivity considerations due to using proximity in cluster assignment, the family of density based clustering algorithms pioneered by DBSCAN~\cite{ester1996density,schubert2017dbscan} makes local neighborhood a prime consideration in forming clusters. However, the density-based clustering family does not offer a convenient prototype for each cluster, and is thus limited in its applicability to human-in-the-loop pipelines such as those outlined in~\cite{repfairness}. In particular, density-based clusterings could yield non-convex clusters, where the centroid computed over cluster objects could be situated outside the natural boundaries of the cluster. Our method, as we will see, will leverage concepts from local neighborhood assessments from the density-based clustering family, and use that within the framework of centroid clustering inspired by $K$-Means. 

\subsection{Quantifying Local Connectivity}\label{sec:quantloc}

Local connectivity in cluster membership desert, as illustrated in the previous section, can be thought of as: {\it how well the local neighborhood of the data object supports its membership to the cluster in question}. We now consider quantifying local connectivity at the object level, which will be aggregated to the level of different clusters in order to arrive at a measure of how well local connectivity is adhered to, in a given clustering. This quantification would form an evaluation metric for assessing local connectivity in clustering. 

Consider an object whose cluster-specific local neighborhood is conceptualized as the set of its $t$ nearest neighbors (we use $t$ instead of the conventional $k$ to avoid conflict with the $K$ in $K$-Means) within the cluster in question. We would like the $t$ nearest neighbors to comprise objects that:

\begin{itemize}[leftmargin=*]
    \item {\bf Offer a Cluster Pull:} We would like the neighbors to offer a pull in the direction towards the cluster prototype. While {\it pull} is admittedly an informal word, we believe it is fairly straightforward to interpret the meaning. To illustrate this notion, observe that the local neighborhood in Fig~\ref{fig:motivation}(a) was largely red objects which may be seen as pulling the object towards the red cluster. This is in sharp contrast with the local neighborhood pull towards the blue cluster in Fig~\ref{fig:motivation}(b). 
    \item {\bf Are Proximal to the Object:} Even if the $t$ nearest neighbors are towards the cluster prototype and can be seen as offering a pull, such a pull is meaningless unless the neighbors are proximal to the object in question. For example, consider Fig~\ref{fig:motivation}(c) where the neighbors of the black object are all towards the blue cluster. However, the appeal of this pull is dented by the fact that the neighbors are quite distant from the black object. 
\end{itemize}

We now quantify the above desired characteristics in the form of a quantitative measure, for a given clustering. Let $X$ be the data object in question, and $C$ be the cluster prototype to whom the local connectivity strength is to be estimated. The dataset of objects involved in the clustering is denoted as $\mathcal{X}$. Given our interest in quantifying the pull towards the cluster prototype, we first identify the set of $t$ nearest neighbors of $X$ that are both: (i) members of the cluster in question i.e., $C$, and (ii) lie {\it in between} $X$ and the cluster prototype for $C$. This set is denoted as $N_t^C(X)$:

\begin{equation}
    N_t^C(X) = \mathop{\arg\min}_{S \subseteq C \wedge Satisfies(S,X,C) \wedge |S| = t} \sum_{s \in S} dist(s,X)
\end{equation}

where:

\begin{multline}
    Satisfies(S,X,C) = \\ \bigwedge\limits_{s \in S} (dist(s,C) \leq dist(X,C)) \wedge (dist(X,s) < dist(X,C))
\end{multline}

$Satisfies(.,.,.)$ enforces the condition that objects in $N_t^C(X)$ fall in between $C$ and $X$ through a distance check; the first distance condition checks whether each element $s$ is closer to $C$, and the second checks whether it is on the {\it 'same side'} of $C$ as $X$ is. Among objects that satisfy these conditions, $t$ of them that are most proximal to $X$ are chosen to form the set $N_t^C(X)$. It may be noted that in cases where there are not enough objects that satisfy the eligibility condition, $|N_t^C(X)|$ may be less than $t$. This is likely to happen when $C$ is very close to $X$; we will outline its implications later. 

\begin{figure}
  \includegraphics[width=0.4\columnwidth]{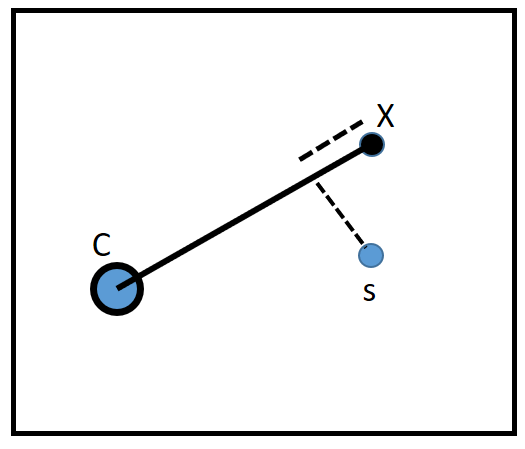}
  \caption{Quantifying Local Conenctivity Illustration}
  \label{fig:geometric}
\end{figure}

Our interest is now in assessing how well objects in $N_t^C(X)$ adhere to the {\it pull} and {\it proximity} heuristics outlined above. We use a simple geometric intuition in order to quantify these. Consider Figure~\ref{fig:geometric} where $X$ is the black object and $C$ is the big blue encircled object, as before. The small blue object is $s \in N_t^C(X)$. Consider the line joining $X$ and $C$ and $s$ shown as being projected on to the line. The {\it pull} heuristic would prefer the dotted line indicating the projection of $s$ to the line to be {\it as short as possible} since that would direct the pull offered by $s$ to be aligned towards $C$. The {\it proximity} heuristic, on the other hand, would prefer $s$ to be as close as possible to $X$, thus preferring that both the dotted lines be as short as possible. We would additionally like the local connectivity to be comparable across different data objects in $\mathcal{X}$. Thus, we measure the two distances indirectly in relation to the distance between $X$ and $C$, as two measures, {\it Deviation} (Dev) and {\it Normalized Distance} (ND), as follows:

\begin{equation}
    Dev(X,C,s) = \frac{dist(C,s) + dist(s,X)}{dist(X,C)} - 1.0
\end{equation}

\begin{equation}
    ND(X,C,s) = \frac{dist(X,s)}{dist(C,s) + dist(s,X)}
\end{equation}

$Dev(X,C,s)$ would evaluate to $0.0$ when $s$ falls directly on the line connecting $X$ and $C$, since that would ensure that $dist(C,s) + dist(s,X) = dist(X,C)$. $Dev(X,C,s)$ increases the more $s$ deviates from that line, leading to its name. $ND(X,C,s)$ on the other hand, measures the distance between $X$ and $s$ as a fraction of the distance between $X$ and $C$ through $s$. Thus, $ND(.,.,.)$, unlike $Dev(.,.,.)$ is directly related to the length of both dotted lines in Fig~\ref{fig:geometric}. Since we would like both of these measures to be numerically small ($\approx 0$), we would like to minimize the product of these, which we call as the {\it local connectivity disagreement} measure:

\begin{equation}
    LCD(X,C,s) = Dev(X,C,s) \times ND(X,C,s)
\end{equation}

Higher values of $LCD()$ denote lower levels of {\it local connectivity} offered by $s$ to support the membership desert for the pair $X,C$. This disagreement may be aggregated across all objects in $N_t^C(X)$ to arrive at an object level estimate:

\begin{equation}
    LCD(X,C) = \sum_{s \in N_t^C(X)} LCD(X,C,s)
\end{equation}

When $|N_t^C(X)|<t$, the $LCD$ would be correspondingly lower since there are fewer objects to sum over. Since we expect $|N_t^C(X)|<t$ to happen when $X$ is already very close to $C$, this translates to an alternative route to reduce $LCD$ for such objects; in addition to improving local connectivity by way of neighbors' positions, {\it $LCD$ can also be improved (i.e., numerically reduced) through enhanced proximity between objects and their cluster prototypes, which would lead to smaller $|N_t^C(X)|$}. Among objects in the cluster $C$, some may have high LCDs and some may have lower values for LCD. Towards assessing a cluster, consider using the average of the LCDs across all objects as an aggregate measure. This would enable a small set of objects with very shallow local connectivity (i.e., high LCD scores since LCD measures disagreement) to be ignored due to being compensated by a large number of low LCD scores across other objects in the cluster. This may be considered undesirable in the face of the high importance accorded to the concern for the most disadvantaged, such as in the very popular stream of Rawlsian notions of fairness~\cite{john1971theory}. Motivated by such considerations, we accord the cluster with an LCD value computed as the highest LCD (i.e., lowest local connectivity, since LCD measures disagreement) among its objects:

\begin{equation}
    LCD(C) = \mathop{\max}_{X \in C} LCD(X,C)
\end{equation}

A clustering of a dataset would produce multiple clusters, since a clustering defines a partitioning. In order to arrive at a dataset-level measure of connectivity offered by a clustering, we would need an aggregate statistic at the dataset level. As in the case above, we would like to ensure that no cluster suffers from bad local connectivity, making the highest LCD among clusters a natural measure to minimize. We call this {\it MaxLCD}. Additionally, We would also like to minimize LCD across all clusters, making {\it AvgLCD} a very pertinent measure. 

\begin{equation}
    MaxLCD(\mathcal{C}) = \mathop{\max}_{C \in \mathcal{C}} LCD(C) \\
\end{equation}

\begin{equation}
    AvgLCD(\mathcal{C}) = \frac{1}{|\mathcal{C}|} \sum_{C \in \mathcal{C}} LCD(C)
\end{equation}

These are analogous to the construction of Max Wasserstein and Avg Wasserstein used in evaluation of fair clustering~\cite{abraham2020fairness}. Thus, {\it MaxLCD} and {\it AvgLCD} offer quantifications of disagreement with local connectivity across the dataset, as manifested in the clustering $\mathcal{C}$. A good clustering would be one which, in addition to performing well on traditional clustering evaluation metrics such as {\it purity} and {\it silhoutte}, achieves {\it low values of MaxLCD and AvgLCD} (thus, high local connectivity). 

\subsection{Drawbacks of LCD Measures}

While LCD measures are, we believe, a starting point for quantifying local connectivity, these are not free of shortcomings. We outline a few drawbacks, which could potentially point to ways of refining them to yield better metrics of local connectivity. 

{\it First}, both $Dev(.)$ and $ND(.)$, which form the building blocks of LCD measures, rely on distances expressed as fractions of other distances. This makes them unable to be sensitive to variations in absolute distances. Consider the case of $Dev(.)$; when $X$ and $C$ are close to each other, even slight deviations of $s$ from the straight line connecting them are amplified, with $dist(X,C)$ forming the denominator. Similarly, take the case of $ND(.)$; high values of $dist(C,s)$ push it towards $0.0$ by providing a very high denominator. When $X$ and $s$ are very far from $C$, even high values of $dist(X,s)$ could cause $ND(.) \approx 0$. Such cases make $LCD$ less meaningful to quantify the connectivity of fringe objects that are far from cluster prototypes. Any attempts at addressing such absolute distance issues should also care to retain the comparability of the resultant metrics across objects in the dataset. {\it Second}, we have excluded neighbors of $X$ that do {\it not} belong to the same cluster as $C$, from consideration in $N_t^C(X)$. This means that an object's neighbors' pull towards the assigned cluster is evaluated without regard to whether it has similar or stronger pulls towards other clusters. This, we believe, is a minor issue, since such stronger pulls towards a different cluster also would likely reduce cluster coherence in general. This means that any clustering that attempts to improve coherence of clusters in addition to local connectivity (such as our method, $LOFKM$, introduced later) would address this implicitly to some extent using the cluster coherence criterion. 

The above two sets of drawbacks are not meant to be comprehensive but to serve to provide a flavour of the possibilities of improving upon LCD measures, and the challenges in those directions. 


\section{LOFKM: Enhancing Local Connectivity in Clustering}\label{sec:lofkm}

We have argued and motivated that local connectivity is a crucial factor in considering membership desert for an object to a cluster. Local neighborhood statistics has been extensively used in the stream of work on density-based clustering, initiated through the popular DBSCAN clustering method~\cite{ester1996density}. Density-based clustering has the ability to identify clusters that have non-convex shapes (e.g., can disambiguate star and crescent\footnote{https://en.wikipedia.org/wiki/Star\_and\_crescent} as separate clusters) and overlapping convex shapes (e.g., can identify rings arranged concentrically as separate clusters). However, this ability comes at a cost; density-based clustering inherently lacks the possibility of choosing a meaningful representative prototype for a cluster (e.g., in the above cases, observe that the centroid would lie outside the cluster itself and would be meaningless as a prototype). Our method, LOFKM, is the result of an attempt to bring a density-based flavour within $K$-Means framework, in order to improve local conncectivity considerations. 

Our design considerations are as follows:

\begin{itemize}[leftmargin=*]
\item {\it Conceptual Simplicity:} We would like to retain the conceptual simplicity inherent in $K$-Means which has likely been at the core of it's widespread popularity. Additionally, we would like to bring in density-based concepts within it in a lucid manner. 
\item {\it Computational Convenience:} The task of clustering is a dataset-level optimization problem which has inherent complexities. This makes directly using local connectivity measures (e.g., LCD) within the optimization infeasible. Due to solving a computational task, computational convenience is also a significant consideration. 
\end{itemize}

\subsection{Towards a Method}

As we have seen, {\it local connectivity} involves a relation between an object and a cluster prototype in the backdrop of the local neighborhood of the object {\it in the 'direction' of the cluster prototype}. It is important to note that the local neighborhood of an object is a property of its location within the similarity space provided by the pre-specified $[dataset, distance\ function]$ pair, and is in no way {\it 'alterable'} to nudge clustering towards deepening local connectivity (or any other consideration, for that matter). 

High $LOF$ (Ref. Sec~\ref{sec:localneighborhood}) objects are more likely to suffer from shallow local connectivity since their neighborhood is sparse; so the neighbors are unlikely to support their membership to any cluster by much. One way to enhance local connectivity would be through better {\it inlineness}, which would be to set cluster prototypes in such directions from high LOF objects within which they have many neighbors. This, however, would require a significantly different prototype construction, putting the conceptual simplicity of $K$-Means prototype estimation at risk. Yet another way would be to bring the cluster prototype towards such high $LOF$ objects, which would enhance their connectivity through both support from neighborhood as well as lower $|N_t^C(X)|$. This route is amenable to exploration while staying within the framework of the $K$-Means clustering formulation, and forms the basis of our {\it LOFKM} method. However, it risks bringing down the compactness of the cluster, which is a factor that would have repercussions on other metrics such as cluster purity and silhoutte as well. As obvious, deepening a particular normative consideration in any machine learning task is expected to introduce constraints that would reduce the clustering quality overall; in other words, higher local connectivity is not expected to come {\it 'for free'}. A good clustering under the local connectivity lens would be one that can deepen local connectivity with {\it limited impact} on other metrics of clustering quality; this, we will see, is the focus of our empirical evaluation. 

\subsection{LOFKM: The Method}

In line with the idea of bringing cluster prototypes closer to higher LOF data objects, we start with assigning a weight to each data object, as follows:

\begin{equation}\label{eq:lofweight}
    W(X) = 
    \begin{cases}
    1.0 & LOF(X) \leq 1 \\
    LOF(X) & otherwise
    \end{cases}
\end{equation}

$W(X)$ is simply the LOF score bounded under by $1.0$. This weight is then used in re-formulating the standard $K$-Means objective as follows, for a given clustering $\mathcal{C}$ over the dataset:

\begin{equation}\label{eq:objective}
    \sum_{C \in \mathcal{C}} \sum_{X \in C} W(X) \times \bigg( \sum_{A \in \mathcal{A}} (X.A - C.A)^2 \bigg)
\end{equation}

where $A$ is any attribute from the set of attributes $\mathcal{A}$, with $X.A$ and $C.A$ denoting the value taken for the attribute by the object $X$ and the cluster prototype of cluster $C$ respectively (notice that we have overloaded $C$ to denote both the cluster and its prototype for notational simplicity). Intuitively, this is equivalent to considering the dataset as comprising each object as being replicated as many times as its LOF score requires, and applying standard $K$-Means over the enlarged dataset. There are two sets of variables that we can change in order to optimize for the objective; the {\it cluster memberships} and {\it cluster prototypes}. Standard $K$-Means optimizes these in turn (keeping one set fixed, and optimizing for the other) over many iterations until the cluster memberships stabilize. 

Under the objective in Eq~\ref{eq:objective}, the membership assignment step, given the cluster prototypes, is as follows:

\begin{equation}\label{eq:estep}
    \forall\ X \in \mathcal{X}, \ \ Cluster(X) = \mathop{\arg\min}_{C \in \mathcal{C}} \sum_{A \in \mathcal{A}} (X.A - C.A)^2
\end{equation}

Since we are updating each object independently given the current estimate of cluster prototypes, $W(X)$ does not factor into this cluster assignment step since it is simply a constant factor for each $X$ independent of which cluster $X$ gets assigned to. This, as one may notice, is {\it exactly the cluster assignment step in $K$-Means}. It may sound odd as to why we critique the $K$-Means membership desert and still use it in {\it LOFKM}; the crucial factor here is that this proximity-based membership desert is used against a set of cluster prototypes that are estimated in very sharp contrast to the analogous step in $K$-Means. The {\it LOFKM} cluster prototype estimation step is as follows:

\begin{equation}\label{eq:mstep}
    \forall\ A \in \mathcal{A}, \ \ C.A = \frac{\sum_{X \in C} W(X) \times X.A}{\sum_{X \in C} W(X)}
\end{equation}

In other words, each $X$ is accounted for as many times as warranted by $W(X)$. 

Towards generating a clustering from a dataset, much like in $K$-Means clustering, we start with a random initialization of cluster prototypes followed by iteratively applying Eq~\ref{eq:estep} and Eq~\ref{eq:mstep} until the cluster memberships become relatively stationary across iterations. Owing to these steps mirroring those of standard $K$-Means, we do not outline a full pseudocode for {\it LOFKM} herewith. 

\subsubsection{Note on Complexity}

The $K$-Means steps, much like the usual $K$-Means algorithm, is linear in the number of objects, number of clusters and number of attributes. However, computing the weights, i.e., Eq~\ref{eq:lofweight}, is more expensive. While LOF computation is generally regarded as between superlinear and quadratic in the number of objects~\cite{breunig2000lof}, faster methods have recently been proposed~\cite{lee2016fast,babaei2019detecting}. It is notable that any further advancements in improving LOF computations readily transfer over to LOFKM as well, given that the LOF and $K$-Means steps are decoupled within LOFKM.

\begin{table}[tbp]
\begin{tabular}{|c|c|c|c|}
\hline
{\bf Name} & {\bf \# Instances} & {\bf \# Attributes} & {\bf \# Classes} \\
\hline
Yeast & 1484 & 8 & 10 \\
\hline
Wireless\footnote{short for Wireless Indoor Localization} & 2000 & 7 & 4 \\
\hline
Avila & 20867 & 10 & 12 \\
\hline
\end{tabular}
\caption{Dataset Statistics}
\label{tab:dataset}
\end{table}

\section{Experimental Evaluation}\label{sec:expts}

We now describe our empirical evaluation. We start by outlining the datasets and baselines in our empirical evaluation, while also outlining the evaluation setup. This is followed by detailed results from empirical evaluation and analyses. 

\subsection{Datasets, Baselines and Evaluation Setup}

\subsubsection{Datasets}

We evaluate our methods on multiple real-world datasets from the UCI Machine Learning Repository. These have widely different numbers of objects, ranging from $1.5k$ to $21k$, and spread across $4-12$ classes. The dataset statistics are summarized in Table~\ref{tab:dataset}. 

\subsubsection{Baseline}

Much like the only two existing papers that propose new normative considerations in clustering, that of proportionality~\cite{chen2019proportionally} and representativity~\cite{repfairness}, we use the classical $K$-Means formulation as the baseline method in our experimental evaluation. We do not include either of the above methods in our comparison since they optimize for significantly different notions of membership desert; as an example, it may be seen that the method from~\cite{chen2019proportionally} was used in the empirical evaluation for representativity in~\cite{repfairness}, and it was observed (unsurprisingly) that the basic $K$-Means fared much better than~\cite{chen2019proportionally} on representativity. 

\subsubsection{Evaluation Setup}

We follow the evaluation framework for fair clustering (as in~\cite{abraham2020fairness,repfairness}), with the evaluation being conducted across two kinds of metrics; (i) {\it local connectivity} (analogous to fairness metrics in fair clustering) metrics, viz., {\it AvgLCD} and {\it MaxLCD}, and (ii) clustering quality metrics, viz., {\it silhoutte~\cite{rousseeuw1987silhouettes}} (Sil) and {\it clustering purity\footnote{https://nlp.stanford.edu/IR-book/html/htmledition/evaluation-of-clustering-1.html}} (Pur). For {\it LOFKM}, we expect improvements on the former, and setbacks on the latter. {\it LOFKM} may be judged to be effective if it is able to achieve good gains on the former set of metrics, at reasonable detriment to the latter. For both LOFKM and $K$-Means, we average the performance metrics across $100$ random starts, so as to achieve stable and reliable numbers. We always set the number of clusters in the output, i.e., the parameter $K$, to be equal to the number of classes in the respective datasets (Ref. Table~\ref{tab:dataset}). 


\begin{table*}[tbp]
\begin{tabular}{|c|c|c|c|c|c|c|c|}
\hline
\hline
{\bf Dataset} & {\bf Method} & \multicolumn{3}{c|}{\it AvgLCD $\downarrow$} & \multicolumn{3}{c|}{\it MaxLCD $\downarrow$} \\
\cline{3-8}
& & {\it t = 3} & {\it t = 4} & {\it t = 5} & {\it t = 3} & {\it t = 4} & {\it t = 5} \\
\hline
\hline
\multirow{3}{*}{Yeast} & KM & 0.93 & 1.18 & 1.42 & 1.20 & 1.57 & 2.00 \\
                       & LOFKM & 0.92 & 1.08 & 1.15 & 1.15 & 1.53 & 1.98 \\
                       \cline{2-8}
                       & Improvement \% & 01.07\% & 08.47\% & 09.01\% & 04.17\% & 02.55\% & 01.00\% \\
\hline
\multirow{3}{*}{Wireless} & KM & 1.24 & 1.68 & 2.04 & 1.32 & 1.79 & 2.24 \\
                       & LOFKM & 1.18 & 1.56 & 1.90 & 1.31 & 1.73 & 1.95 \\
                       \cline{2-8}
                       & Improvement \% & 04.83\% & 07.14\% & 06.87\% & 00.76\% & 03.35\% & 12.95\% \\
\hline
\multirow{3}{*}{Avila} & KM & 1.11 & 1.48 & 1.83 & 1.33 & 1.77 & 2.19 \\
                       & LOFKM & 0.99 & 1.31 & 1.61 & 1.32 & 1.80 & 2.19 \\
                       \cline{2-8}
                       & Improvement \% & 10.81\% & 11.49\% & 12.02\% & 00.75\% & -01.69\% & 00.00\% \\
\hline
\hline
\multicolumn{2}{|c|}{Avg of Improvement \%} & 05.57\% & 09.03\% & 09.30\% & 01.89\% & 01.40\% & 04.65\% \\
\hline
\hline
\end{tabular}
\caption{Evaluation on Local Connectivity Measures. Note that lower values are better for both {\it AvgLCD} and {\it MaxLCD}, as indicated using the arrow in the column heading.}
\label{tab:lcd}
\end{table*}

\begin{table*}[tbp]
\begin{tabular}{|c|c|c|c|c|c|c|c|}
\hline
\hline
{\bf Dataset} & {\bf Method} & \multicolumn{3}{c|}{\it Sil $\uparrow$} & \multicolumn{3}{c|}{\it Pur $\uparrow$} \\
\cline{3-8}
& & {\it t = 3} & {\it t = 4} & {\it t = 5} & {\it t = 3} & {\it t = 4} & {\it t = 5} \\
\hline
\hline
\multirow{3}{*}{Yeast} & KM & \multicolumn{3}{c|}{0.26} & \multicolumn{3}{c|}{0.42} \\
\cline{3-8}
                       & LOFKM & 0.27 & 0.27 & 0.26 & 0.41 & 0.41 & 0.41 \\
                       \cline{2-8}
                       & Change \% & +03.84\% & +03.84\% & 00.00\% & -02.40\% & -02.40\% & -02.40\% \\
\hline
\multirow{3}{*}{Wireless} & KM & \multicolumn{3}{c|}{0.40} & \multicolumn{3}{c|}{0.93} \\
\cline{3-8}
                       & LOFKM & 0.39 & 0.39 & 0.39 & 0.77 & 0.78 & 0.78 \\
                       \cline{2-8}
                       & Change \% & -02.50\% & -02.50\% & -02.50\% & -17.20\% & -16.13\% & -16.13\% \\
\hline
\multirow{3}{*}{Avila} & KM & \multicolumn{3}{c|}{0.15} & \multicolumn{3}{c|}{0.46} \\
\cline{3-8}
                       & LOFKM & 0.18 & 0.18 & 0.18 & 0.45 & 0.45 & 0.45 \\
                       \cline{2-8}
                       & Change \% & 20.00\% & 20.00\% & 20.00\% & -02.17\% & -02.17\% & -02.17\% \\
\hline
\hline
\multicolumn{2}{|c|}{Avg of Change \%} & 07.11\% & 07.11\% & 05.83\% & -07.26\% & -06.90\% & -06.90\% \\
\hline
\hline
\end{tabular}
\caption{Evaluation on Clustering Quality Measures. Note that higher values are better for both {\it Sil} and {\it Pur}, as indicated using the arrow in the column heading.}
\label{tab:quality}
\end{table*}

\subsection{Experimental Results and Analysis}

We first outline the structure of the experimental analysis. Local connectivity, as outlined in Sec~\ref{sec:quantloc}, is assessed using a parameter $t$, the number of relevant neighbors for an object; this parameter is used in the computation of both {\it MaxLCD} and {\it AvgLCD}. For {\it LOFKM}, there is a similar parameter in the input, which is the number of neighbors for an object used in LOF computation (Ref. Sec.~\ref{sec:localneighborhood}). These, being similar in spirit, are set to identical numbers, and we denote both as $t$. We experiment with varying values of $t$; in the interest of brevity, we report results for $t \in \{3,4,5\}$ as a representative set of results since the trends held good for higher values. $KM$, short for $K$-Means, does not use any neighborhood parameter in the method. 

The evaluation on fairness metrics is illustrated in Table~\ref{tab:lcd} whereas the evaluation on clustering quality appears in Table~\ref{tab:quality}. The percentage change of the {\it LOFKM} metric over that in {\it KM} is indicated explicitly, for ease of interpretation. An average of $5-10\%$ gains are achieved on the {\it AvgLCD} measure, indicating a sizeable improvement in local connectivity in the clusterings output by {\it LOFKM} over those of {\it KM}. Further, the improvements are seen to improve with the size of the dataset, which is expected since larger datasets allow for more flexibility in clustering assignments. The corresponding improvements in {\it MaxLCD} are seen to be smaller. {\it MaxLCD} quantifies the worst local connectivity across clusters, and thus relates to the quantification over a single cluster, which in turn is the worst local connectivity across members of the cluster. While it would intuitively be expected that least locally connected objects which would be in sparse regions where local connectivity improvements would be harder to achieve, it is promising to note that {\it LOFKM} consistently achieves improvements on {\it MaxLCD} over {\it Yeast} and {\it Wireless}; the corresponding improvements in {\it Avila} are limited, and negative in one case. The trends on the clustering quality metrics in Table~\ref{tab:quality} may be regarded as quite interesting. It may be noted that $t$ does not play a role for results of {\it KM} since the clustering quality metrics as well as $KM$ are agnostic to $t$. As outlined earlier, we expect that the cost of local connectivity enhancement in {\it LOFKM} would manifest as a deterioration in clustering quality. While we can observe such deterioration in {\it Pur} in Table~\ref{tab:quality}, {\it LOFKM} is surprisingly able to achieve improvements in {\it Sil} on the {\it Yeast} and {\it Avila} datasets. On careful investigation, we found evidence to hypothesize that {\it LOFKM} discovers {\it secondary clustering structures}, which differ from the primary ones that are better correlated with external labels ({\it Pur}, as one might remember, measures correlation with external labels). These secondary clustering structures, while not necessarily tighter, are found to be well separated, yielding improvements in {\it Sil}. This interestingly correlates with similar observations over {\it Sil} in representativity fairness (Ref. Sec. 6.3.2 in~\cite{repfairness}). In contrast to {\it Yeast} and {\it Avila}, {\it Wireless} does not seem to exhibit such well-separated secondary structures, leading to falls in both {\it Sil} and {\it Pur}. Across datasets, the deterioration in {\it Pur} is seen to be fairly limited, to within $10\%$; we would re-iterate that the fact that such deterioration comes with an improvement in {\it Sil} indicates the promisingness of {\it LOFKM}. To summarize, {\it LOFKM} is seen to offer consistent and often sizeable improvements in local connectivity, with mixed trends in clustering quality. 

\section{Discussion}\label{sec:discussion}

Having considered local connectivity as a factor for membership desert in clustering, it is useful to think about how this relates to other notions and other factors that may be argued to play a role in membership desert. 

Local connectivity is distinctly different from {\it representativity}~\cite{repfairness} in that an object that is very distant from the cluster prototype could still be locally connected to the very same cluster. While this conceptual distinction cannot be more apparent, in practice, we expect peripheral/fringe objects of a cluster to suffer from local connectivity, and similar could be true for representativity as well. In a way, local connectivity provides a way to distinguish between objects in the periphery of clusters that are locally connected to the cluster and those that are not. This points to the possibility of using both in tandem. Peripheral objects 'deserve' better representativity, but local connectivity could provide a way to prioritize among them.  The connection with {\it proportionality}~\cite{chen2019proportionally} is somewhat more nuanced, since proportionality violations are evaluated at the collection level. That said, proportionality violations may be expected to be in the gulf between existing clusters, since those would be the locations where one would expect to see preference to the existing cluster assignment waning. Thus, addressing proportionality violations by changing cluster assignments may be seen as automatically addressing local connectivity, since the objects would be better locally connected to the new cluster. These relationships between concepts could lead to interesting future explorations. 

Membership desert having been considered along lines of {\it proximity}~\cite{repfairness}, {\it collective vote}~\cite{chen2019proportionally} and {\it local connectivity}, it is interesting to think of whether there are other ways of thinking about cluster memberships. The building blocks of Silhoutte~\cite{rousseeuw1987silhouettes} provide an interesting angle to the issue. Silhoutte quantifies the average distance to the objects of it's existing cluster, and those to the objects of the {\it next nearest} cluster, and uses these to compute a normalized difference, called the object-specific silhoutte co-efficient. The silhoutte score is then the mean\footnote{https://scikit-learn.org/stable/modules/generated/sklearn.metrics.silhouette\_score.html} of these. It may be argued that each object needs to be accorded a minimum level of higher proximity to the existing cluster than the next best, or that objects need to score similarly on their respective silhoutte co-efficients. This line of exploration requires low variance of the silhoutte co-efficients over the dataset, as well as maximizing the minimum silhoutte co-efficient. Another perspective is to consider the role of sensitive attributes such as race, sex, gender and religion, when clustering person-level data. Each of the notions of membership desert could be extended using the role of sensitive attributes. For example, there could be two routes to enhance membership desert based on the relationship with the cluster prototype. One could be through proximity, and another could be through similarity in sensitive attribute profiles, and these could compensate slightly for each other.  This discussion hopefully serves to indicate that there is plentiful meaningful room for enhancing the diversity of membership desert notions in clustering formulation. A recent position paper~\cite{whitherfc} considers certain other normative possibilities within the task of clustering. 

\section{Conclusions and Future Work}\label{sec:conclusions}

In this paper, we investigated, for the first time, local connectivity and its relevance to membership desert in centroid clustering. Through a critique of cluster membership desert focusing on $K$-Means, we motivated the need to consider local connectivity as a crucial normative consideration in deciding cluster memberships, in addition to centroid proximity (the only criterion in classical formulations such as $K$-Means). Following upon this argument, we outlined ways of quantifying local connectivity for a given clustering, to aid evaluating clusterings on the local connectivity criterion. Towards developing a clustering that would promote local connectivity, We considered local neighborhood assessments from the family of density-based clustering methods, and adopted {\it LOF} for usage within the $K$-Means formulation, leading to a local connectivity-oriented clustering method, {\it LOFKM}. Through an evaluation of {\it LOFKM} vis-a-vis $K$-Means (following the evaluation frameworks in similar works~\cite{chen2019proportionally,repfairness}), we illustrated that {\it LOFKM} is able to deepen local connectivity in clustering outputs while producing well-separated clusters at only reasonably degradations to clustering purity as measured against external labels. 


\noindent{\bf Future Work:} We are considering various layers of interplay between local connectivity and notions of fairness as espoused within popular schools such as {\it Rawlsian fairness}~\cite{john1971theory}. Second, we are considering blending local connectivity along with the other normative principles explored in clustering, such as representativity and proportionality. Third, we are considering other criteria for membership desert involving sensitive attribute classes such as gender and ethnicity. Further, we have also been considering the relationships between clustering interpretability (e.g.,~\cite{balachandran2012interpretable}) and fairness. 


\bibliographystyle{ACM-Reference-Format}
\bibliography{repfairness}

\end{document}